\documentclass[twocolumn]{article}

\usepackage{arxiv}

\usepackage[utf8]{inputenc} 
\usepackage[T1]{fontenc}    
\usepackage{hyperref}       
\usepackage{url}            
\usepackage{booktabs}       
\usepackage{amsfonts}       
\usepackage{nicefrac}       
\usepackage{microtype}      
\usepackage{lipsum}		
\usepackage{graphicx}
\usepackage{natbib}
\usepackage{doi}
\usepackage{multicol}
\usepackage{booktabs}
\usepackage{tabularx}
\usepackage{cuted}
\newcolumntype{Y}{>{\centering\arraybackslash}X}

\graphicspath{{Fig/}}

\title{Hespi: A pipeline for automatically detecting information from hebarium specimen sheets}

\author{ \href{https://orcid.org/0000-0003-1274-6750}{\includegraphics[scale=0.06]{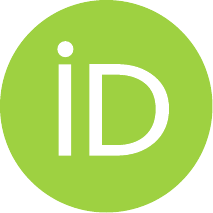}\hspace{1mm}Robert ~Turnbull} \\
	Melbourne Data analytics Platform\\
	The University of Melbourne\\
	Parkville, VIC 3053 \\
	\texttt{robert.turnbull@unimelb.edu.au} \\
	\And
	\href{https://orcid.org/0000-0002-4991-946X}{\includegraphics[scale=0.06]{orcid.pdf}\hspace{1mm}Emily Fitzgerald} \\
	Melbourne Data analytics Platform\\
	The University of Melbourne\\
	Parkville, VIC 3053 \\
	\texttt{emily.fitzgerald@unimelb.edu.au} \\
	\And
	\href{https://orcid.org/0000-0002-5498-0556}{\includegraphics[scale=0.06]{orcid.pdf}\hspace{1mm}Karen Thompson} \\
	Melbourne Data analytics Platform\\
	The University of Melbourne\\
	Parkville, VIC 3053 \\
	\texttt{karen.thompson@unimelb.edu.au} \\
	\And
	\href{https://orcid.org/0000-0002-8226-6085}{\includegraphics[scale=0.06]{orcid.pdf}\hspace{1mm}Joanne L. Birch} \\
	School of BioSciences\\
	The University of Melbourne\\
	Parkville, VIC 3053 \\
	\texttt{joanne.birch@unimelb.edu.au} \\
}

\renewcommand{\headeright}{Preprint}
\renewcommand{\undertitle}{Preprint}
\renewcommand{\shorttitle}{Hespi}

\hypersetup{
pdftitle={Hespi},
pdfauthor={Robert ~Turnbull},
}

\begin{document}

\begin{strip}
  \centering
  \maketitle
  \vskip\baselineskip

\begin{abstract}

Specimen-associated biodiversity data are crucial for biological, environmental, and conservation sciences. A rate shift is needed to extract data from specimen images efficiently, moving beyond human-mediated transcription. We developed `Hespi' (HErbarium Specimen sheet PIpeline) using advanced computer vision techniques to extract pre-catalogue data from primary specimen labels on herbarium specimens. Hespi integrates two object detection models: one for detecting the components of the sheet and another for fields on the primary primary specimen label. It classifies labels as printed, typed, handwritten, or mixed and uses Optical Character Recognition (OCR) and Handwritten Text Recognition (HTR) for extraction. The text is then corrected against authoritative taxon databases and refined using a multimodal Large Language Model (LLM). Hespi accurately detects and extracts text from specimen sheets across international herbaria, and its modular design allows users to train and integrate custom models.

\end{abstract}

\noindent\makebox[\textwidth]{}
  \vskip\baselineskip


\end{strip}

\section{Introduction}\label{intro}

There are an estimated 3,500 active herbaria globally, containing approximately 398 million physical specimens \citep{TheirsIH}. Specimen-associated biodiversity data are sought after for biological, environmental, climate, and conservation sciences \citep{Lacey2017, Davis2023}. Increased investment in natural history collections-based digitization efforts has significantly increased the availability of high-resolution specimen images in the last decade \citep{Davis2023, Walker2022}. However, transcription rates associated with the digitization of biodiversity data have remained flat \citep{Vollmar2010,Guralnick2024}. A rate shift for data extraction from specimen images is required to eliminate this impediment to the mobilization of biodiversity data. Advanced computer vision techniques hold the potential to achieve that rate shift by reducing the time required for digitization of text-based label data and, in doing so, to mobilize vast quantities of biodiversity data from digital specimen images \citep{thompson2023_mapping, Walker2022}. Natural Language Processing methods also have the potential to increase accuracy of text digitization from specimen labels by enabling language detection and terminology extraction \citep{Owen2020}. This paper introduces `Hespi' a HErbarium Specimen sheet PIpeline. Hespi detects components of specimen sheets, detects the fields in the primary specimen labels and recognizes the text using Optical Character Recognition (OCR), Handwritten Text Recognition (HTR) and multimodal Large Language Models (LLMs).  

\section{Background}\label{background}

Herbarium specimens contain a preserved biological sample and both primary (e.g. taxonomic identity, collector, collection date or location) and secondary (e.g. redetermination or confirmation of taxonomic identity, date of curation event) collection data. These specimens provide a verifiable record of the presence of a taxon at a point in time \citep{Funk2018, Kirchhoff2018}. Historically, specimen-associated data were documented on paper \citep{Groom2019, Walton2022}; recorded in field notebooks, transcribed into printed catalogs, and primary and secondary data were written on labels that were attached to the specimens. Mobilization of these specimen data is typically achieved by processing specimens through a digitization workflow, involving the production of a digital specimen image followed by the extraction of text data from that digital image either manually (i.e., via a human intermediary) or semi-automatically \citep{thompson2023_mapping, Hidalga2020, Kirchhoff2018, Nelson2015}. Their digitization, conforming to biological data standards [e.g., ABCD (Access to Biological Collections Data) \citep{Holetschek2012} and DarwinCore \citep{Wieczorek2012}], is essential for maintaining their accuracy and ensuring their availability for reuse \citep{Groom2019}. The conversion of imaged labels into digital text and the parsing of that text into standard data fields are some of the slowest steps in the digitization pipeline and are significant bottlenecks for biodiversity data mobilization \citep{Groom2019,Guralnick2024, Walton2022, tulig2012increasing, Kirchhoff2018}.

Deep learning models using artificial neural networks have shown to be effective for data extraction from digital images, including phenological \citep{Pearson2020, MoraCross2022} or morphological \citep{MoraCross2022, Wilson2022} data, taxonomic identifications \citep{shirai2022}, and other textual data \citep{Walker2022, Guralnick2024, Milleville2023}. Such techniques hold potential to reduce the reliance on human-mediated transcription and processing of specimen data \citep{Owen2020, Milleville2023}. Neural network models require large datasets of carefully curated and labeled images for training, to create models that achieve good performance \citep{Walker2022}. The optimal techniques and parameters for tasks required for data extraction from herbarium specimens continue to be elucidated. We have previously described training a deep learning model to detect the various components of a specimen sheet \citep{thompson2023_identification}. Here we extend this approach as part of a larger pipeline to extract textual information from primary specimen labels to enable their digitization.

 Optical Character Recognition (OCR) protocols have long been recognized as holding potential for mobilization of textual data from specimens. However, this potential has not yet been realized due to limitations in the accuracy of data extraction using OCR software. Workflows have progressed from applying OCR to whole specimen sheets \citep{drinkwater2014use, haston2012developing, tulig2012increasing}, to (manually) identifying the label area and applying OCR \citep{alzuru2016cooperative, anglin2013improving, barber2013salix, dillen2019benchmark, haston2015}. It has been demonstrated that applying OCR to the label-only image is more effective than applying OCR to the whole image \citep{alzuru2016cooperative,haston2015,owen2019d4}, and that running OCR over individual text lines cropped from a label image is faster than processing the whole label \citep{owen2019d4}. 

The capture of handwritten text is one of the most challenging aspects of optical character recognition. Handwritten text does not always conform to standard character shapes or sizes, which poses a significant challenge for OCR \citep{Owen2020}. The text on natural history specimen labels provides additional challenges, as labels may contain a mixture of handwritten and typed text or the handwriting of multiple individuals \citep{Owen2020}. Tests of accuracy for software targeted at Handwritten Text Recognition (HTR) (e.g., ABBYY Fine Reader Engine, Google Cloud Vision) indicate that HTR tools can capture a proportion of specimen data that is accurate and of high quality suggesting that this technology is already a viable technique for data capture \citep{Owen2020, haston2015}. 

Large Language Models (LLMs) are trained to predict the next token in a textual sequence. This self-supervised task allows for massive datasets to be used in training, and produces models which are able to perform sophisticated tasks in natural language processing. As \cite{Radford2019LanguageMA} state: `high-capacity models trained to maximize the likelihood of a sufficiently varied text corpus begin to learn how to perform a surprising amount of tasks without the need for explicit supervision.' Multimodal LLMs allow non-textual data inputs such as images and combine these sources to handle complex data processing tasks.  Recently, multimodal LLMs have been applied to the field of document understanding \citep{borchmann2024notesapplicabilitygpt4document}. It was found that multimodal LLMs show significant promise for document analysis and the results are vastly improved when including results from OCR in the input to the LLM. The application of multimodal LLMs to the extraction of text from specimen sheets has been discussed by \citeauthor{weaver2023} and \citeauthor{Guralnick2024}.

\section{Material and Methods}\label{methods}

\begin{figure*}[ht!]
\includegraphics[width=\textwidth]{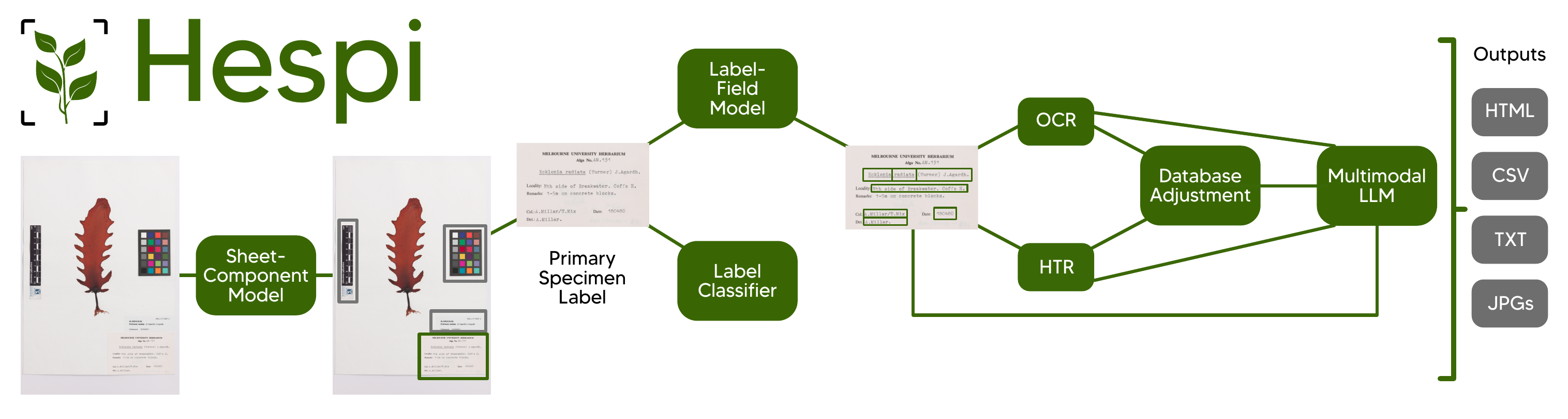}
\caption{The Hespi pipeline. Specimen sheet available at \url{https://online.herbarium.unimelb.edu.au/collectionobject/MELUA118997a}.}
\label{fig:pipeline}
\end{figure*}

A diagram of the Hespi pipeline is shown in Fig. \ref{fig:pipeline}. The stages of the pipeline are explained in more detail below. Briefly, Hespi first takes a specimen sheet and detects the various components within it using the Sheet-Component Model. Any full primary specimen label that is detected is cropped and serves as input for the Label-Field Model, which detects text in a subset of data fields written on the label. A neural network Label Classifier is used to determine the type of text (typeset or handwritten) on the label. The text within each field is recognized using OCR and HTR engines. The recognized text is post-processed and cross-checked against authoritative plant and fungal name lists for specific fields. The primary specimen label and the recognized text are given to a multimodal LLM for correction. The final extracted labels and text data are written to an HTML report and a CSV file for viewing and subsequent data processing. 

\subsection{The Sheet-Component Model}
\label{sec:sheet-component}

\begin{figure*}
\includegraphics[width=\linewidth]{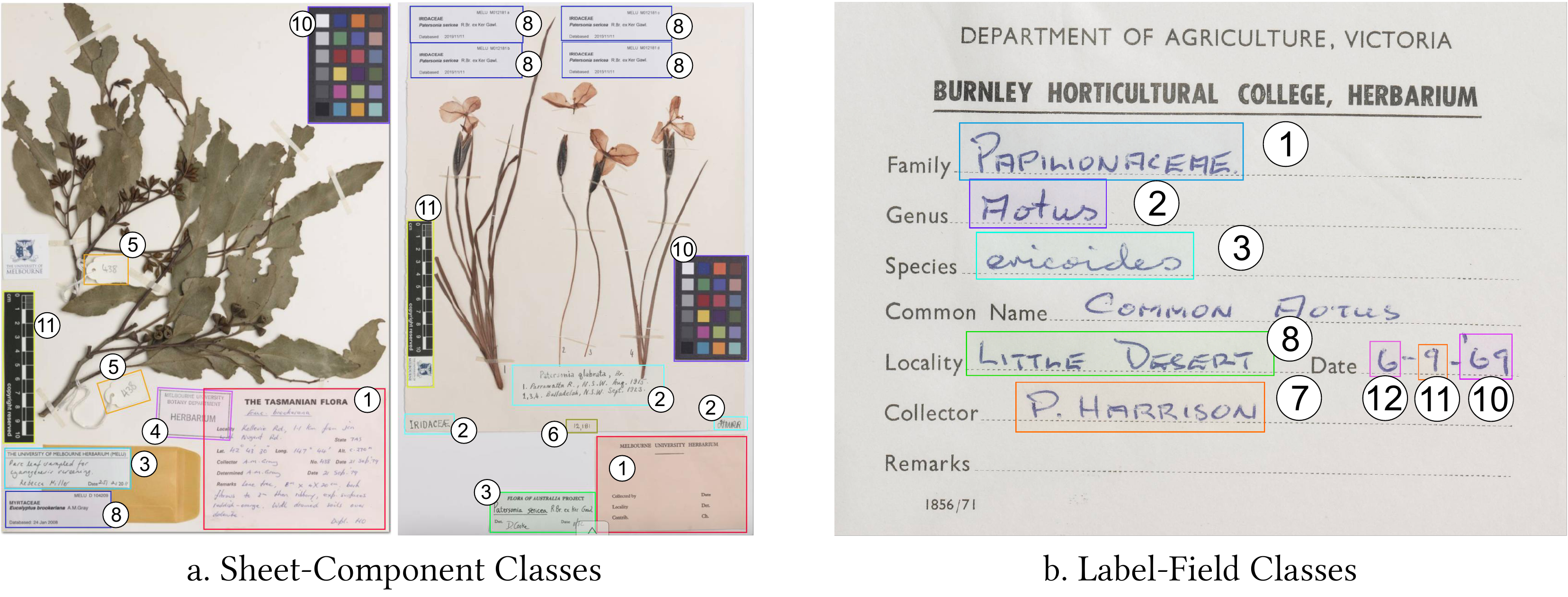}
\caption{Examples of the prediction classes for the Sheet-Component model (a) and the Label-Field model (b). Numbers correspond to listing of classes in the respective sections. Specimen sheets available at  \url{https://online.herbarium.unimelb.edu.au/collectionobject/MELUD104209a} and  \url{https://online.herbarium.unimelb.edu.au/collectionobject/MELUD111492a}.}
\label{fig:classes}  
\end{figure*}

This object detection model is based on the work of \cite{thompson_data_2023}. It takes specimen sheet images and outputs bounding boxes for 11 components:
\begin{enumerate}
    \item Primary specimen label
    \item Data on the specimen sheet outside of a label (`original data'; often handwritten)
    \item Taxon and other annotation labels
    \item Stamps
    \item Swing tags attached to specimens
    \item Accession number (when outside the primary specimen label)
    \item Small database labels
    \item Medium database labels
    \item Full database labels
    \item Swatch
    \item Scale
\end{enumerate}

Examples of these components are shown in Fig. \ref{fig:classes}a. A dataset with annotations corresponding to these components is publicly available on FigShare \citep{thompson_data_2023}. It includes 4,821 specimen sheet images annotated with bounding boxes for the various components (see `Sheet-Component Model' above), of which 1180 are designated for validation. Of these, 4,371 come from The University of Melbourne Herbarium (MELU) and 50 images come from each of the following nine herbaria (with acronyms as per \cite{TheirsIH}) from the benchmark dataset by \cite{dillen2019benchmark}:

\begin{itemize}
\item Meise Botanic Garden (BR)
\item Royal Botanic Gardens, Kew (K)
\item The Natural History Museum, London (BM)
\item ZE Botanischer Garten und Botanisches Museum, Freie Universität Berlin (B)
\item Royal Botanic Garden Edinburgh (E)
\item Muséum National d'Histoire Naturelle, Paris (P)
\item University of Tartu (TU)
\item Naturalis Biodiversity Center (L)
\item University of Helsinki (H)
\end{itemize}

\begin{figure*}[htbp!]
\includegraphics[width=\textwidth]{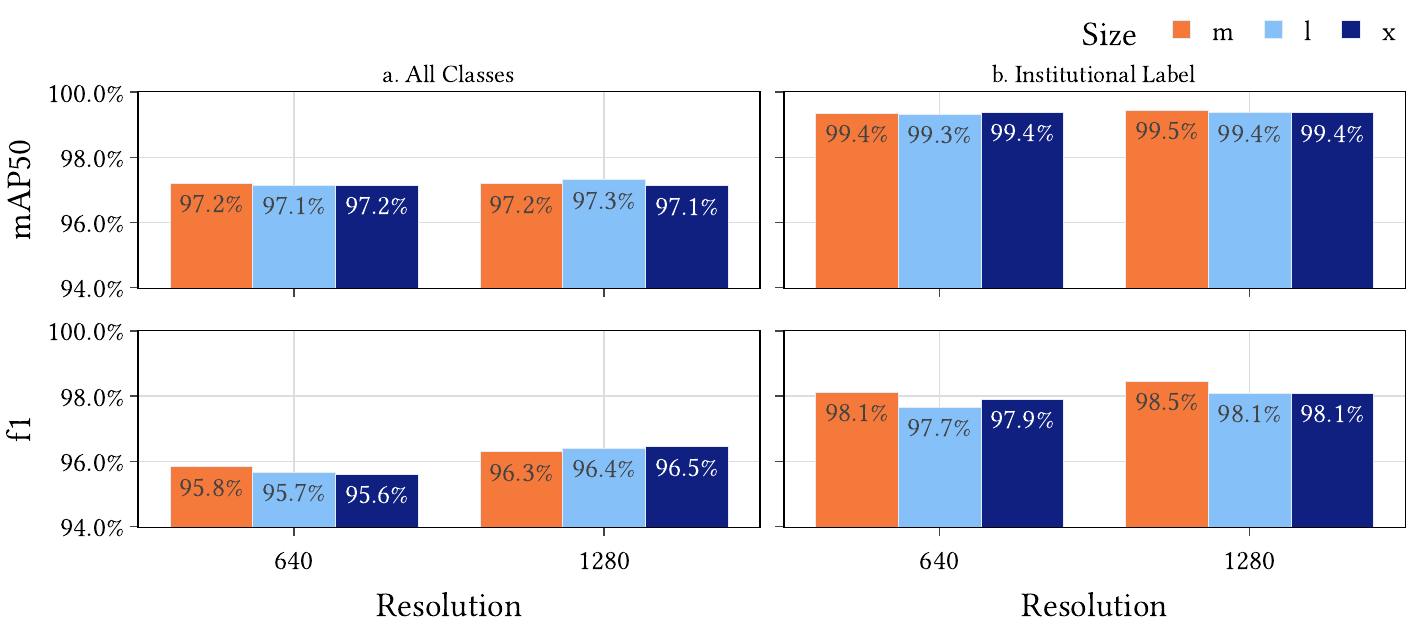}
\caption{Validation results for YOLOv8 models trained on the Sheet-Component annotations. The left column shows the results for all component classes and the right column shows the result for just the primary specimen label.}
\label{fig:Sheet-Component-Result-mAP-f1}
\end{figure*}

\cite{thompson2023_identification} discuss the results for training a version of this model using YOLOv5 (You Only Look Once) \citep{YOLOv5}. YOLO is a neural network object detection model which simultaneously predicts bounding boxes and class probabilities. Thompson et al. trained a YOLOv5 model on the MELU annotations and then fine-tuned on the annotations for the other nine herbaria. Here we use YOLOv8 \citep{YOLOv8} and train using all the annotations together. Six versions of the model were trained, with resolutions at 640 and 1240 pixels and sizes `m', `l' and `x'. The mean average precision at an intersection over union value of 50\% (mAP50) and the f1 score on the validation set for each of these models is shown in Fig. \ref{fig:Sheet-Component-Result-mAP-f1}. The most critical component for Hespi is the `primary specimen label', which refers to the label generated when the specimen is first accessioned into an herbarium. The `primary specimen label' typically contains information about the specimen that needs to be digitized and this is the component used for downstream tasks in the pipeline. The highest f1 score for accurate detection of the primary specimen label was 98.5\%, which was achieved with a model size of `x' and a resolution of 1280. This is the configuration of the model that is used in the Hespi pipeline.

\subsection{The Label-Field Model}
\label{sec:label-field}

The Label-Field Model takes any primary specimen label detected from the Sheet-Component Model and detects bounding boxes for the following fields:

\begin{enumerate}
    \item Family
    \item Genus
    \item Species
    \item Infraspecific taxon 
    \item Authority of the taxon at the lowest rank provided 
    \item Collector's field number
    \item Collector
    \item Locality
    \item Geolocation (latitude, longitude, elevation, and elevation units)
    \item Year
    \item Month
    \item Day
\end{enumerate}

Examples of these components are shown in Fig. \ref{fig:classes}b.

\begin{figure*}[htbp]
\includegraphics[width=\textwidth]{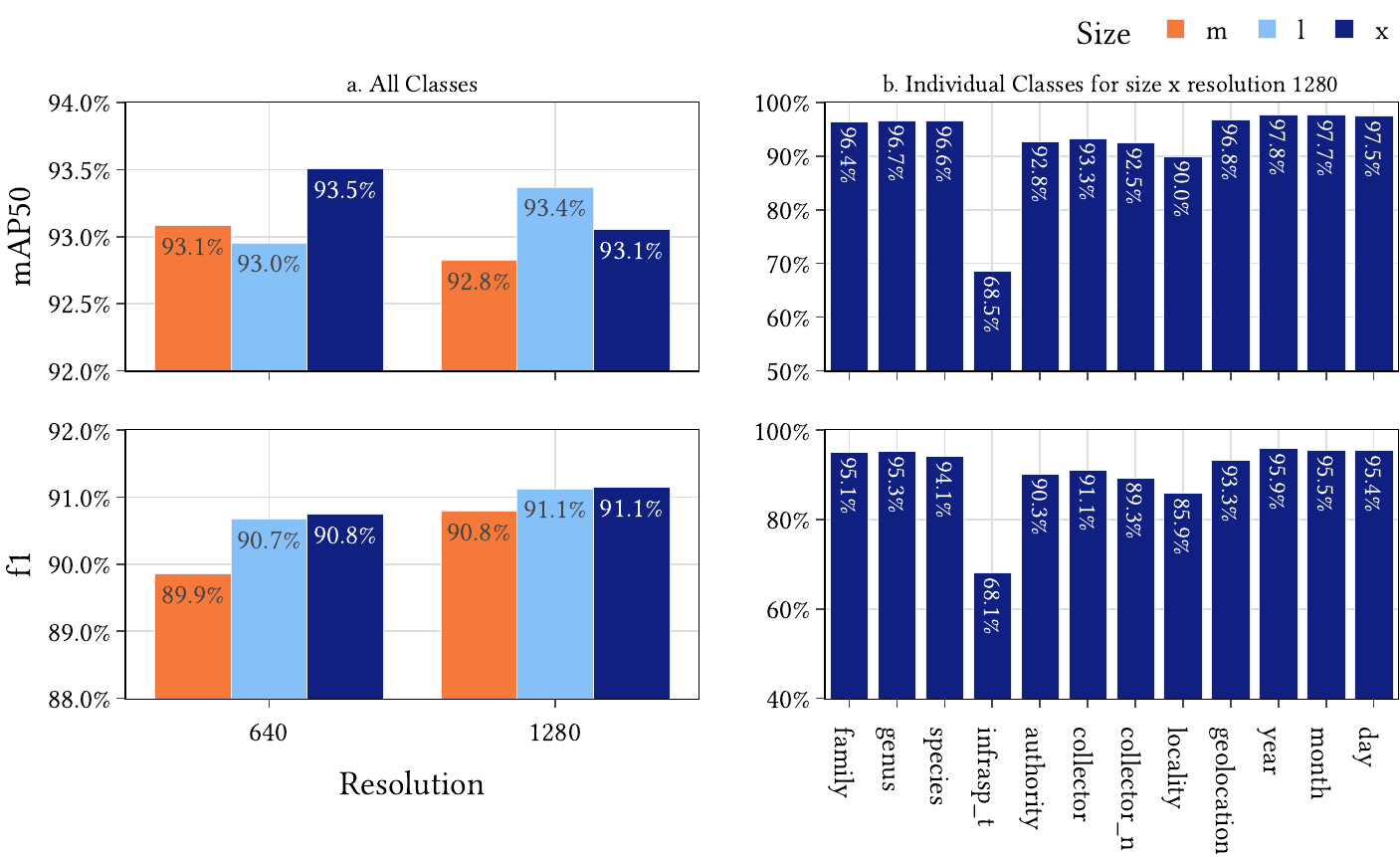}
\caption{The results on the validation set for the Label-Field model. The left column (a) shows the results for all classes at different resolutions and YOLOv8 model sizes. The right column (b) shows results for each individual class at a model size of `x' and a resolution of 1280.}
\label{fig:label-field-results}
\end{figure*}

These classes were annotated on 3,642 images of primary specimen labels from 10 herbaria. A total of 2,603 images are from The University of Melbourne Herbarium and the remainder are from the nine herbaria represented in the benchmark dataset described by \citep{dillen2019benchmark}. These were broken down into 2,887 training images and 755 validation images. The images and annotations are available on FigShare \citep{label_field2024}. The model was trained using YOLOv8 at three different sizes: `m', `l' and `x'; and at two resolutions: 640 and 1280 (Fig. \ref{fig:label-field-results}a.) The model size `x' at a resolution of 1280 gave the highest overall f1 score and this is the model configuration that is used in the Hespi pipeline. The results of this model for each field class are shown in Fig. \ref{fig:label-field-results}b.

\subsection{Label Classifier}
\label{sec:label-classifier}

We have trained a classifier to detect the following types of writing on the primary specimen label:
\begin{itemize}
    \item typewritten
    \item printed
    \item handwritten
    \item combination
    \item empty
\end{itemize}

\begin{figure}[htbp]
\includegraphics[width=\columnwidth]{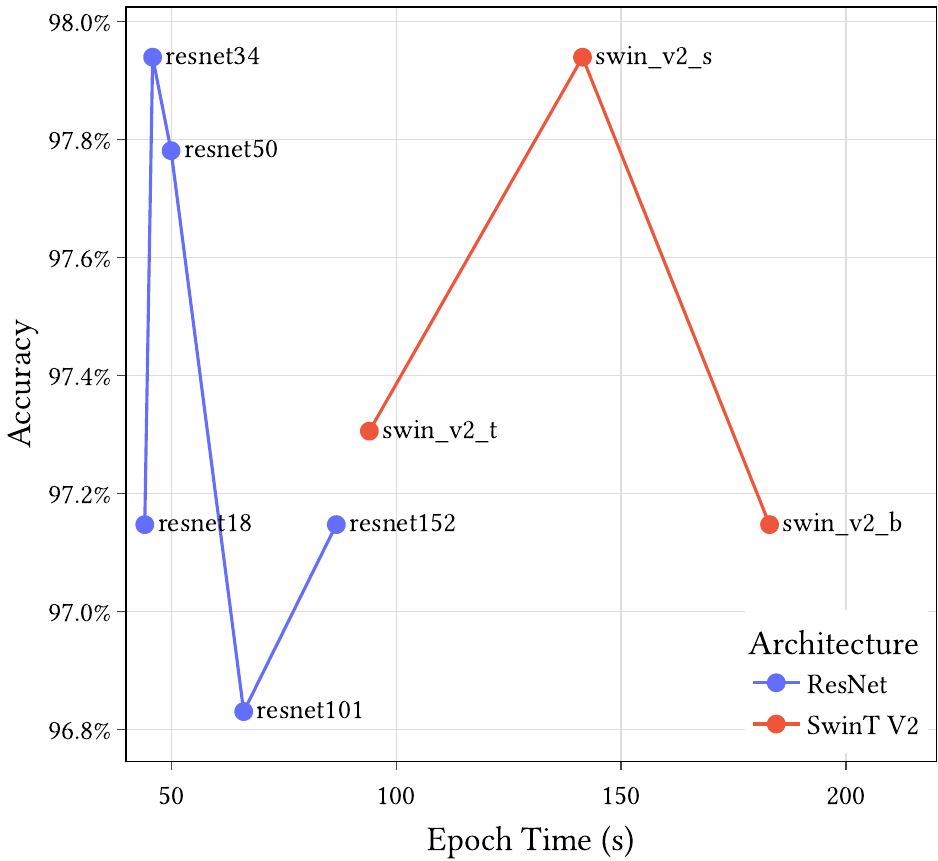}
\caption{Validation results compared with training time per epoch for the label classifier models.}
\label{fig:classifier-results}
\end{figure}

These writing types were annotated to the 3,152 images from the MELU dataset. This dataset was partitioned into 2,521 training images and 631 validation images. Images and annotations are available on FigShare \citep{classifier2024}. Pretrained ResNet \citep{resnet} and Swin Transformer V2 \citep{swin_v2} models were used and fine-tuned on this dataset using torchapp \citep{torchapp} for 20 epochs with a batch size of 16 and at a resolution of 1024 pixels. The best performing models were the ResNet-34 and the Swin Transformer V2 of size `s' with an accuracy of 97.9\% (Fig. \ref{fig:classifier-results}). The ResNet-34 model is used as part of the Hespi pipeline due to its lower computational complexity.

\subsection{Text Recognition}

Each field detected by the Label-Field Model is input into the Text Recognition module. This uses the Tesseract OCR engine \citep{tesseract2007,TessOverview} and the TrOCR `large' HTR model \citep{trocr}. 

Text formatting is applied to Tesseract/TrOCR results for the family, genus, and species fields, where a standardized format is expected. Family and genus fields are changed to title case, species to lower case. For all three, punctuation marks are stripped from the beginning and end of the text, as well as any whitespace/empty characters. 

For the family, genus, species and authority fields, any recognized text is cross-checked against the World Flora Online database \citep{WFO}, an international compendium of vascular plants and mosses, and against databases within the Australian National Species List \citep{aunsl}, a nationally recognized taxonomy of Australian biodata \citep{Cooper2023}. The AuNSL databases used by Hespi are the:
\begin{itemize}
    \item Australian Plant Name Index \citep{APNI}
    \item Australian Bryophyte Name Index \citep{AusMoss}
    \item Australian Fungi Name Index \citep{AFNI}
    \item Australian Lichen Name Index \citep{ALNI}
    \item Australian Algae Name Index \citep{AANI}
\end{itemize}

If the extracted text matches to a taxonomic name in the reference datasets with a similarity of 80\% or more using the Gestalt (Ratcliff/Obershelp) approach \citep{gestalt}, Hespi will assign the taxonomic name from the reference dataset to that field. In this way, minor differences of the taxon name/s on the specimen label or the extracted data to those in taxonomic reference datasets are corrected. Such differences may be orthographic variants, incorrect spelling of the taxon name on the primary specimen label, or incorrect text recognition. The closeness of the matches indicate to Hespi whether to use the output from Tesseract or TrOCR when recording the text of the other fields. If no text recognition method is found to be superior (i.e., they generate the same score, or both scores fall below 80\%), then handwritten or mixed labels will use the output from TrOCR and other labels will use output from Tesseract.

\subsection{Large Language Model (LLM) Correction}

After the text recognition, the results are passed through a multimodal large language model (LLM) to correct any errors. By default, Hespi uses OpenAI's `gpt-4o' model \citep{openai2023gpt4}. This can be changed to any other model from OpenAI or Anthropic by specifying the model name on the command line. The LLM is prompted with the image of the primary specimen label, the list of the desired fields, the currently accepted text for each field and the outputs from the OCR and HTR engines and how the text has been adjusted after cross-checking with the relevant datasets. The LLM is requested to output the text for any fields where the accepted text is incorrect. Currently, no examples of this process are provided through the prompt and so Hespi is using the LLM as a `zero-shot' learner \citep{Radford2019LanguageMA}. Hespi could be modified to provide the LLM with examples of images from a particular herbarium and so use the LLM as a `few-shot' learner which will likely improve the results for similar primary specimen labels \citep{few_shot}. This is left for future experimentation.

\subsection{Outputs}

\begin{figure}[htbp]
\includegraphics[width=\columnwidth]{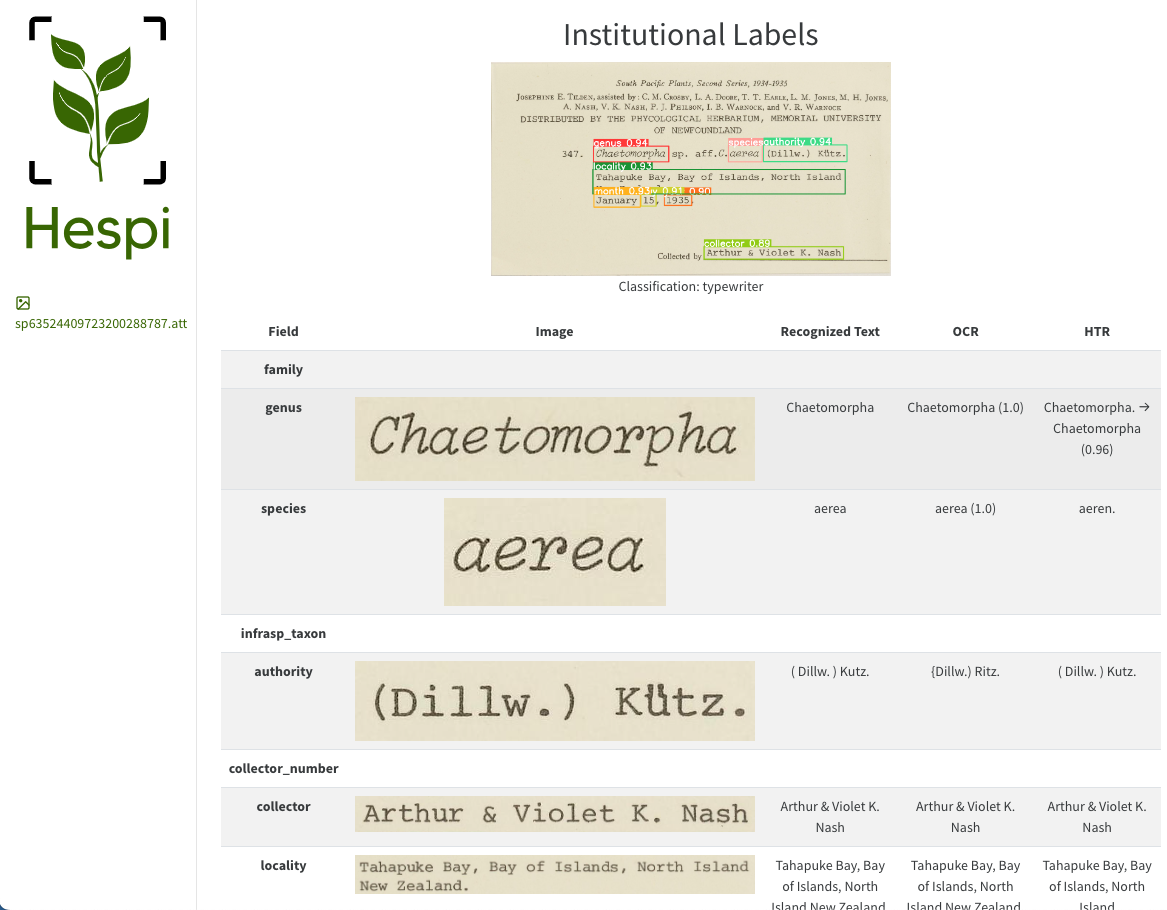}
\caption{An example screenshot of a Hespi HTML report.}
\label{fig:hespi-report}
\end{figure}

Hespi produces a directory of outputs with the cropped image files and the predictions of both the Tesseract and TrOCR results in both CSV and text files. The pipeline outputs are summarized as an HTML report which displays the cropped images from each model and the derived recognized text (Fig. \ref{fig:hespi-report}). In this way it is possible to manually cross-check the accuracy of the derived text by comparing it with the original data, visualized from the entire  specimen label or the corresponding extracted data field. The CSV file includes the match score between 0 and 1 for the family, genus, species, and authority, alongside all OCR and HTR results. These scores are a value between 0 and 1, with 1 indicating a perfect match and no corrections made; 0.8 to 1.0 indicating how similar a match was, and 0 indicating no match found with a similarity of 80\% or higher. 

\section{Accuracy of data extraction on unseen datasets}

We created three test datasets for Hespi to evaluate its performance end-to-end. The first, `MELU-T', consists of 100 images of specimen sheets from The University of Melbourne Herbarium where the primary specimen label was either printed or typewritten. The second, `MELU-H', consists of 100 images from The University of Melbourne Herbarium specimen sheets where the primary specimen label was handwritten. The final dataset, `DILLEN', consists of 100 images from the benchmark dataset by \cite{dillen2019benchmark}, including at least ten specimen sheets from each of the nine institutions represented in that dataset. There were no overlaps between the test datasets and any of the other datasets used for training and validation. Each test dataset includes the classification of the type of text on the label and the text for each field on the primary specimen label, which is subsequently referred to as the `ground truth' text. The test datasets and the evaluation script are available online \citep{testdatasets2024}. The evaluation script shows the accuracy of the label classification and whether or not any particular field should be empty. It also evaluates the similarity between the predicted and the ground truth text for each field. The percentage similarity is measured using the Gestalt (Ratcliff/Obershelp) algorithm. Only fields where text is provided in either the test dataset or the predictions are included in the results. If a field is present in either the test dataset or the predictions but not the other then the similarity is zero. All non-ASCII characters and punctuation are removed and the results are case-insensitive. The aggregate results for predictions from Hespi using various components of the pipeline are shown in Table \ref{table:results}. This includes results with and without the LLM correction.

The distribution of similarity scores for each field type using the full Hespi pipeline is shown in Fig. \ref{fig:test-results}. Plots of distribution of similarity scores without using an LLM are provided in the supplementary material.

The results for the MELU test datasets were higher for all metrics than those of the DILLEN dataset (Table \ref{table:results}). This is unsurprising given that a higher proportion of the training data was from MELU and the DILLEN dataset is highly diverse in both the layout of the primary specimen labels and in the languages in which the label data are recorded. LLM correction improved the mean similarity score substantially across each data set. Fields with the taxonomic classification of the specimen (i.e. family, genus and species) are predicted well across the datasets probably because this information is prominent on each label and due to the post-processing correction of the recognized text from the dataset of known entities.

\begin{table*}[]
  \centering
  
\begin{tabularx}{\textwidth}{YYYYYYYYY}
\toprule
\shortstack{Test \\ Dataset} & \shortstack{Label \\ Classification} & \shortstack{Field \\ Present} & \multicolumn{2}{c}{\shortstack{Text Similarity \\ No LLM}} & \multicolumn{2}{c}{\shortstack{Text Similarity \\ LLM correction}} \\

   & Accuracy & Accuracy & Median & Mean & Median & Mean  \\

\midrule
MELU-T    &    100.0\%    &    98.5\%    &     100.0\%    &    91.4\%    &    100.0\%    &    92.7\% \\
MELU-H    &    99.0\%    &    97.6\%    &      100.0\%    &    81.1\%    &    100.0\%    &    88.8\% \\
DILLEN    &    87.0\%    &    84.9\%    &    75.3\%    &    56.8\%    &    100.0\%    &    67.1\% \\

\bottomrule
\end{tabularx}
\vspace*{0.5em}
\caption{Aggregate results for the three test datasets with text similarity results provided with and without LLM correction.}
\label{table:results}

\end{table*}

\begin{figure*}[htbp]
\includegraphics[width=\textwidth]{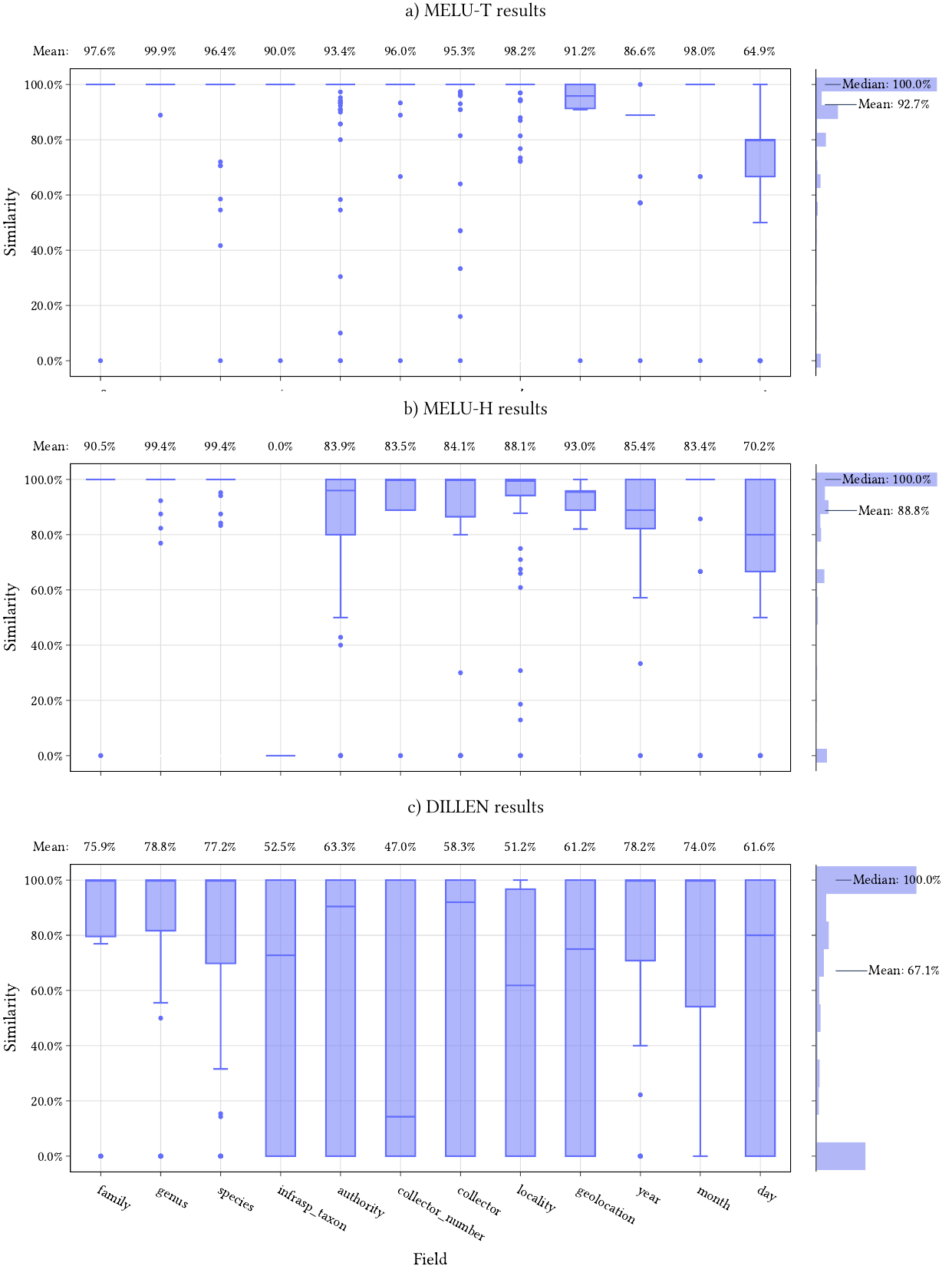}
\caption{Test results for the full Hespi pipeline, including LLM correction. Box plots showing the quartiles of the text similarity scores for all fields found in the primary specimen labels of the three test datasets. Outliers shown as points. A histogram of results for all fields in each dataset shown on right.}
\label{fig:test-results}
\end{figure*}

\section{Discussion}

This study establishes a modular pipeline integrating computer vision and OCR technologies that takes specimen digital image files as input and extracts text data from target fields on the specimen label in a versatile digital file format as output. The pipeline facilitates classification of specimens according to the format of their label data (handwritten, printed/typewritten, or a combination), which enables the scripted collation of specimen sheets with consistent data formats for efficient downstream data handling. Pre-processing of taxonomic identity data is achieved through comparison to national taxonomic names lists, with percentage similarity parameters scores reported in the output files. The text is corrected using a multimodal Large Language Model, which markedly improves the results. Extracted text data are parsed into standard Darwin Core fields and can be visually (manually) checked against the label image in an HTML report that is generated.

\subsection{Modular AI components}
The work presented here achieves significant innovations in the application of deep learning for the digitization of specimen associated biodiversity data: the first, object detection of all major non-plant sample sheet components including rulers, color bars, and text data on labels or handwritten on specimen sheets; secondly, object detection of individual data fields from the primary institutional specimen label; thirdly, the application of specialized Handwritten Text Recognition software; fourthly, the use of multimodal Large Language Models for text correction.

The components of the Hespi pipeline are modular, allowing models to be replaced or fine-tuned to specific data distributions for improved performance. Fine-tuned object-detection models (both Sheet-Components and Label-Field) may easily improve results for objects that are not yet accurately detected (e.g. intraspecific taxon data field) or for previously unseen components (e.g. barcodes). As we investigated previously \citep{thompson2023_identification}, starting with pretrained model weights and fine-tuning as few as 30 new image annotations was sufficient for accurate bounding box detection of known, but previously infrequently encountered, sheet components. Instructions for fine-tuning the open-access model weights are available in Hespi's documentation. The particular LLM used for correction can also be replaced by other LLMs as they are developed.

Primary specimen labels are of primary interest for digitization of specimen data and are the text-bearing image file that is typically input into OCR applications e.g., \citep{barber2013salix, Guralnick2024, Heidorn_labelx}. The Sheet-Component model in Hespi produces a set of images of the extracted primary specimen labels, with file names assigned from the input specimens. The Hespi pipeline detects primary specimen labels with high measures of accuracy (mAP50 values 99.3--99.5\%; f1 values 97.7--98.5\%), including for specimens with labels of different formats and with different placement on the specimens. Extracted images of the primary specimen labels are smaller in size, more efficient for file transfer, and can be viewed more efficiently than digital images of entire specimens. Prior work has shown that OCR performance is improved when only text-bearing labels are submitted for processing \citep{Owen2020}. The execution time of OCR processing is also decreased by restricting the input to only text-bearing images \citep{Kirchhoff2018}. 

\subsection{Sorting of Specimen} 
Digital sorting of specimen labels holds the potential to improve efficiency and accuracy of curation workflows by increasing the format consistency within datasets \citep{Granzow2010, drinkwater2014use, tulig2012increasing, Granzow2010}. For example, Drinkwater et al. \citep{drinkwater2014use} note that more rapid downstream manual data capture and quality control was achieved when specimen sheets were sorted by the country from which specimens were collected or by the collector's name. Collation of similar specimen sheets is enabled in the Hespi pipeline in multiple ways. The Label Classifier enables segregation of labels according to their print type, enabling downstream curation according to requirements of those data types. Additionally, the Hespi pipeline enables rapid and accurate detection and extraction of other text-bearing label types (e.g. annotation labels and data written directly on the specimen sheet) and other sheet components (e.g., rulers and color bars) enabling sorting of these specimens, for efficient downstream curation of a set of specimens with consistent data or data formats.  

The diversity of formats of specimen labels within a single institution over time, and among national and international herbaria, has provided a major challenge for automated digitization of specimen data \citep{tulig2012increasing, Guralnick2024, Owen2020}. The Hespi pipeline has consistently detected 11 of the 12 the target data fields on labels of herbarium specimens drawn from ten international herbaria, which collectively display significant label format diversity. Targeted data fields on the primary specimen labels were consistently and accurately detected (mAP50=90.0--97.8\%; f1 85.9--95.9\%), with the exception of the infraspecific taxon field. The infraspecific taxon data field was poorly predicted in these models (mAP50=68.5\%; f1=68.1\%), presumably due to the rarity of this field appearing in the training data.

\subsection{Detection of textual fields}

Accurate parsing of textual data extracted from entire specimen labels into Darwin Core fields remains a significant challenge \citep{haston2015} and continues to represent an impediment to full automation of the post image capture process. By detecting individual data fields within the primary specimen label, Hespi reduces the need for downstream parsing of textual data. 

Taxon name fields (family, genus, and species) were consistently accurately detected (mAP50=96.4--96.7\%, f1=94.1--95.3\%). These results are comparable to those extracted by Quaesitor \citep{Little}, which achieved 0.80--0.97 recall and a 0.69--0.84 precision in the detection of Latin scientific names in the 16 most common languages for biodiversity articles. The detection of infraspecific taxon names was low (mAP50=68.5, f1=68.1), a challenge that was also noted by \cite{Little}, who noted that infraspecific rank was difficult to discern from hybrid combinations. \cite{Milleville2023} applied Google Cloud Vision API to test accuracy of taxon name recognition from 50 specimen labels, noting that only 24 percent of the words in taxon names were recognized correctly and 36 percent were partially recognized. Extraction of the taxon authority field (mAP50=92.8\%, f1=90.3\%) enabled comparison of extracted taxonomic names with those present in taxon lists during post-processing. Our post-processing results indicate that comparisons with global lists of names (e.g. International plant names index), which include both current names and synonyms, increases accuracy of taxon names where global taxonomic diversity is represented.      

Numerical date fields were readily detected (mAP50=97.5--97.8\%, f1=95.4--95.9\%), although the automated extraction of date data risks failing to differentiate the date/s of specimen collection, determination/s, or receipt in a collection. Therefore post-processing remains necessary to ensure the extracted date data are accurately assigned. Locality data were more challenging as they are provided as free text and, for many of the specimens in the Dillen dataset, were provided in a variety of languages other than English. Georeference data suffered from OCR reading errors, for example, transforming symbols and adding spacing \citep{Guralnick2024} that were, in some cases, able to be eliminated through scripted, standardized corrections.

The Label-Field model prioritises a subset of fields drawn from the `Minimum Information about a Digital Specimen' (MIDS) digitization standard \citep{Haston2022}. For collection curators, the data fields selected hold utility as they collectively serve as a pre-catalogue, representing a feasible and efficient starting point to obtain baseline data about collection holdings \citep{Granzow2010, tulig2012increasing} that can subsequently inform targeted curation efforts. For research that relies on specimen associated data, the extracted data fields are fit for purpose for a broad range of downstream research applications \citep{Haston2020}. 

The data fields selected were those that are typically present on labels in standardized formats, regardless of the institution from which they were accessioned, which potentially increases the likelihood that they can be accurately detected using these model-based methods. Other free text based data fields (e.g. habitat, descriptive or collection notes etc) may be more challenging for bounding-box detection. Those data fields are equally important for extraction to ensure they are available in biodiversity repositories and for subsequent research \citep{Guralnick2024}.

\subsection{Text Recognition}
The Hespi pipeline includes some post-processing of the text recognition, to minimize the need for downstream manual processing. One aspect of this is simply formatting text to nomenclatural protocols, so that fields identified as family or genus are capitalized, and those identified as species or an infraspecific taxon are set as lower case; another was stripping punctuation from the beginning and end of the family, genus, and species fields.

Another aspect is the cross-referencing of text identified as family, genus, species, or authority against a list of each, developed from the Australian National Species List and the World Flora Online Taxonomic Backbone. Initially these reference lists were derived from five indices of the Australian National Species List, incorporating vascular plants (angiosperms, pteridophytes and gymnosperms), bryophytes (mosses, hornworts and liverworts), fungi, lichen, and algae. To extend the cross-referencing ability beyond Australian species, taxonomic information from the World Flora Online, which includes vascular plants (angiosperms, pteridophytes, and gymnosperms) and bryophytes. At this stage, the larger list that incorporates names of fungi, lichen, and algae that do not occur in Australia have not been incorporated into the Hespi reference lists. 

This cross reference acts as an important check of the quality of the transcription by Hespi, as well as aiming to improve the utility for Hespi users. If a match for the transcribed and formatted text is found in the relevant list, no changes are made. If there are minor errors in the transcription, these can be corrected to the family, genus, species, or authority that most closely matches the transcript output. The reference list used is determined by the label-field detection---that is, text identified as a genus name will only be checked against the genus reference list. Importantly, changes will only be made to the output if there is a similarity of 80\% or higher between the detected text and a taxonomic name in the reference lists, set higher than the default cutoff of 60\% to limit any possibility of false matches. In addition, users have the ability to change this cutoff score to be even higher, as well as turning off this matching process entirely. 

An important consideration with changing the output for close matches to the reference datasets is the risk of introducing errors, with Hespi potentially changing a correct OCR result to an incorrect one. This occurred in 1.5\% of all results in the three test datasets when comparing the Hespi result to the ground truth data (with only 13 instances in the 839 results). This was shown in the data as instances where the formatted OCR matched with the ground truth data, but was not recognized when compared to the reference lists and so was changed. Closer inspection described below shows that the instances of Hespi changing an OCR result to an incorrect name is actually 0.4\% (three instances out of 839). 

The vast majority of these results (nine of the 13 instances) occurred with authority fields, which allows for more variation in how names are recorded. The LLM model was able to account for many of these variations, and return the same result as the ground truth. Of those that remained, we found that in seven of these nine authority results, the changed result still referred to the same authority, formatted differently. These differences ranged from as simple as spacing differences (‘A.S. George’ vs ‘A.S.George’) or, more commonly, inclusion of initials (‘Wright \& Ladiges’ vs ‘I.J.Wright \& Ladiges’; ‘Nordensk.’ vs ‘H.Nordensk.’). In the remaining two instances, a different result was returned due to the use of non-standard  abbreviations for the authority on the taxon label which was not matched with the standard abbreviation in the reference list (e.g., ‘Schur.’ instead of ‘Schauer’, so Hespi returned ‘Schau.’; `Gom.' instead of `Gomont', so Hespi returned `Gomb.')

In a small number of instances (four of the 13 instances) Hespi's OCR result matched the ground truth data, but the taxon name was changed through use of the reference datasets and LLM. Of these, three changes (two genus and two species results) corrected a spelling error on the original label: 
\begin{itemize}
    \item \textbf{Ahnfletia} torulosa: a misspelling of \textit{Ahnfeltia}, the Hespi result 
    \item \textbf{Odontitis} verna: a misspelling of \textit{Odontites}, the Hespi result 
     \item Solanum \textbf{sisymbrifolium}: a misspelling of \textit{sisymbriifolium}, the Hespi result 
\end{itemize}
In a single instance, Hespi's OCR result matched the ground truth data, but the taxon name was incorrectly matched using the reference datasets and LLM. In this instance the original label had Photinia \textbf{serulata}, a misspelling of \textit{serrulata}. Hespi identified this as the second closest match to \textit{serulata}, but returned the incorrect closest match, \textit{sertulata}.

Overall, use of the reference datasets and the LLM model resulted in 75\% of 189 results with a close but not exact match to the reference datasets that were corrected by Hespi, reducing downstream processing load. The remaining 25\% did return errors, but all but three of these would have also been errors without this step being included. Cross checking would still be required for these scores, to identify those where the close match returned was not correct, alongside manual correction of those scoring 0. The scoring system can also help prioritise downstream processing needs, with those scoring 1 requiring the least intervention, and those scoring 0 the most. 

Hespi ensures that users can see when and where changes have been made, as well as easily identifying where Hespi has not been able to find a match in these reference lists. Changes are indicated in the console when Hespi is run, and the HTML output provided shows the original transcription (both Tesseract and TrOCR) and any changes made, alongside the selected result and an image of the text. In addition, the CSV output shows all OCR results, and the match score for each relevant field. Downstream users can sort the CSV output by these scores, to easily identify those where no changes were made either because a perfect match was found (a match score of 1), or because no close match was found (a match score of 0), as well as the similarity score (between 0.8 and 1.0) when a change has been made. 

Typically, as was done here, the accuracy of data extracted using OCR or HTR technology is assessed by comparison against data resulting from human transcription of specimen labels. However, this comparison can be challenging as textual data are often formatted or interpreted during manual transcription and may not be presented in exactly the same form as the textual data that were written on the label \citep{Owen2020}. A standard protocol for measuring accuracy of data extracted using manual and automated techniques would be beneficial. Additionally, as new  automated data extraction pathways are developed, an automated, efficient protocol for the comparison of data extraction accuracy is required. These assessment tools could also be applied to develop automated post-capture data correction strategies to optimise data quality and increase the rate of generation of high-quality, accurate digital data.     

\section{Conclusion}

We present Hespi, an open-source pipeline for automatically extracting labels containing textual data from herbarium specimen sheets and recognizing a subset of textual data from the original primary specimen label. It takes specimen sheet images and outputs a suite of formatted information, including the text written on the primary specimen label, which is typically the target data of digitization of specimen sheets. The text is corrected using a multimodal LLM which substantially improves the results of standard OCR and HTR engines. Hespi achieves accurate results on the test datasets. The various components of the model can be fine-tuned for other herbaria to improve results in other contexts. It can be incorporated into a wider strategy of digitizing specimen sheets and thereby making available the wealth of data that are associated with those specimens.

\section{Software and Availability}

Hespi is available as open-source software on Github (\url{https://github.com/rbturnbull/hespi}) under the Apache 2.0 Open Source License. It installed directly from the Python Package Index (\url{https://pypi.org/project/hespi/}). The automated testing as part of the Continuous Integration/Continuous Deployment (CI/CD) pipeline has 100\% code coverage. The whole pipeline runs with a single command and instructions for usage are provided in the online documentation (\url{https://rbturnbull.github.io/hespi/}).

\section{Competing interests}
No competing interest is declared.

\section{Author contributions statement}

R.T. conceived the pipeline. R.T., K.T., E.F. and J.B. created the image annotations. R.T. trained the models. R.T. and E.F. developed the software. R.T., K.T., E.F. and J.B. wrote and reviewed the manuscript. J.B. conceived the research and obtained resources to support the collaboration.   

\section{Acknowledgments}
This research was supported by The University of Melbourne’s Research Computing Services and the Petascale Campus Initiative. 
The authors thank collaborators Niels Klazenga, Heroen Verbruggen, Nunzio Knerr, Noel Faux, Simon Mutch, Babak Shaban, Andrew Drinnan, Michael Bayly and Hannah Turnbull for discussions around aspects of this study.

\bibliographystyle{unsrtnat}
\bibliography{reference}

\end{document}